\pdfoutput=1
\documentclass{article}
\usepackage{spconf,amsmath,graphicx,hyperref}
\usepackage[T1]{fontenc}
\usepackage{xcolor}
\usepackage{float}
\usepackage{multirow}
\usepackage{cite}
\usepackage{subcaption}
\newcommand{\gridw}{0.85in}

\title{Instance-Specific Classifier-Free Guidance for Text-to-Image Diffusion/Flow-Based Editing}
\name{Wajeeh Atrash$^{1}$ and Tom Tirer$^{2}$
}
\address{$^{1}$ Tel Aviv University, $^{2}$ Bar-Ilan University}
\begin{document}
\maketitle
\begin{abstract}
In recent years, a variety of prompt-based image editing methods have been developed based on text-to-image (T2I) diffusion/flow models.
Image editing introduces an inherent trade-off between adhering to the original source image and realizing the change specified by the target prompt, which are typically measured by the LPIPS and CLIP-T metrics, respectively. 
Intuitively, this trade-off can be controlled by the classifier-free guidance (CFG) scale. 
Most existing editing methods, though, use a fixed CFG scale, ignoring that different edits require varying levels of modification.
In this paper, we first explore the potential benefits of using an instance-specific CFG 
by exhaustively re-running prominent editing methods with a wide range of CFG scale values on benchmark datasets.
We show that instance-specific setting can significantly improve LPIPS without compromising CLIP-T. 
Yet, tuning based on repeatedly executing an editing pipeline is inefficient.
To prevent slow inference, we train a light-weight neural network that 
predicts the metrics of the underlying editing method for a given edit instance and CFG scale.
At inference time, these predictions enable fast selection of an instance-specific CFG scale. 
Experiments across multiple editing methods and benchmark settings show the performance advantage of our approach over the baseline methods.

\end{abstract}
\section{Introduction}
\label{sec:intro}

In recent years, diffusion \cite{ho2020ddpm,song2021scorebased} and flow-matching \cite{lipman2023flowmatching,esser2024sd3} models have emerged as the dominant generative models for images, capturing a strong image prior that can also be leveraged for editing and restoration \cite{meng2021sdedit,abu2022adir,garber2024image,garber2025zero}.
Conditioning these models on text prompts, enables high-quality text-to-image (T2I) generation \cite{rombach2022high,Saharia2022-bu}. %
To improve generation quality and alignment with the prompt, classifier-free guidance (CFG) \cite{ho2022cfg} has become a widely used technique. It uses the direction between conditional and unconditional (null-prompt) predictions for steering the sampling trajectory. 
Follow-up works  introduce different flavors of CFG \cite{gao2026c2fg,malarz2025betacfg,wang2026cfgctrl,sadat2024apg}, or modify its nature from extrapolation to interpolation \cite{chung2024cfgpp}. Beyond standard CFG-guided generation, the T2I backbones have also been repurposed for image editing, either by manipulating internal attention or noise representations\cite{hertz2023prompttoprompt,huberman2024editfriendly}, by instruction-tuning the model directly\cite{brooks2023instructpix2pix}, or, more recently, by constructing an editing ODE directly within the flow-matching framework\cite{kulikov2025flowedit}.

Using the T2I backbone for image editing introduces an inherent trade-off between preserving the original source image (e.g., preserving object identity or background) and applying prompt-guided modifications. We want the model to change and edit only the parts mentioned in the prompt, while keeping the rest of the image unchanged or with minimal changes. Besides visual comparison between the source and target images, this trade-off is captured mainly by two metrics: (1) CLIP-T \cite{hessel2021clipscore} (higher is better), the cosine similarity between the edited image and the target prompt in CLIP's\cite{radford2021clip} joint embedding space, which stands in for how well the edit was realized, and (2) LPIPS \cite{zhang2018lpips} (lower is better), a learned perceptual distance between the source and edited image that stands in for fidelity, computed over the whole image. %

Editing methods may balance the trade-off using CFG. Usually, each method has its own default fixed CFG scale. This does not take into account that different images and edits need different levels of modification, which implies different optimal CFG scale values. 
Images that require large changes are expected to need a larger CFG scale, which gives a better CLIP-T score but also alters the image more, resulting in a worse LPIPS score.
On the other hand, exaggerating with the CFG scale for a modest edit, may negligibly improve the CLIP-T score while unnecessarily degrade the LPIPS score.

The idea that CFG need not be applied identically throughout sampling has been explored in several works. C\textsuperscript{2}FG \cite{gao2026c2fg} aligns the guidance strength with the diffusion dynamics through a time-dependent decay function, $\beta$-CFG \cite{malarz2025betacfg} shapes the guidance strength over time using a beta-distribution curve, CFG-Ctrl \cite{wang2026cfgctrl} treats the CFG update as a control problem and steers it dynamically along the trajectory, and APG \cite{sadat2024apg} decomposes and rescales the guidance update at each step to curb oversaturation. What all four have in common, though, is that the CFG scale tuning strategy is fixed in advance and shared across every image and prompt a method is run on. 
They also require access to intermediate diffusion steps, which may not always be available for users of an editing system.

In this work, we explore the potential benefits of using an instance-specific CFG scale in editing methods and propose an efficient automatic mechanism for doing so. We first confirm, by exhaustively re-running two prominent editing methods, DDPM inversion\cite{huberman2024editfriendly} and FlowEdit\cite{kulikov2025flowedit}, across a wide range of CFG scale values on benchmark datasets, that setting the scale differently for each editing instance is beneficial. For example, we show that it
can significantly improve LPIPS without compromising CLIP-T. This tuning strategy, however, is time-consuming and computationally expensive. %

To avoid that cost at inference time, we introduce a lightweight neural network that, given a source image, a target prompt, a candidate CFG scale, and the editing method's metrics at its default CFG scale as an anchor point, 
predicts the LPIPS and CLIP-T scores that would be achieved using the candidate CFG scale. At inference time, this let us scan a range of CFG scale values and apply the editing method with the selected instance-specific CFG scale using at most two editing activations, which is a fraction of the cost needed for exhaustive multi-execution tuning. Testing our approach against the default fixed CFG scale for DDPM inversion and FlowEdit shows significant improvement in terms of LPIPS (and qualitative adherence to the source image) with only small effect on the CLIP-T (adherence to the target prompt). 

\section{Background}

\subsection{T2I models and CFG}

CFG, first introduced in\cite{ho2022cfg}, has become an irreplaceable mechanism in T2I synthesis and editing. CFG operates by training a denoising model to work both conditionally and unconditionally, dropping the condition during training with some probability. This allows the model to learn to predict the added noise at timestep $t$ both conditionally and unconditionally, $\epsilon_\theta (z_t,c)$ and $\epsilon_\theta (z_t,\emptyset)$. Using this, CFG allows us to control the tradeoff between sample fidelity and conditional alignment in diffusion models, by calculating what we call the guided noise:
\begin{equation*}
    \tilde{\epsilon}_\theta(z_t, c) = \epsilon_\theta (z_t,\emptyset) + s\left(\epsilon_\theta (z_t,c) - \epsilon_\theta (z_t,\emptyset)\right)
\end{equation*}
where $s \geq 1$ is the CFG guidance scale hyperparameter. Different scale values result in different behavior in the final editing result, 
where a larger CFG scale amplifies the conditioning signal and its influence. Intuitively, a large $s$ is thus expected to increase the CLIP-based similarity between the target prompt and the edited image compared to weaker guided sampling, which reflects more semantically consistent results and stronger edit coherence. This is likely to come at the cost of fidelity to the source image, as reflected by a higher LPIPS score. On the other hand, applying overly large CFG scale values can steer the sampling away from the source image, beyond what is actually necessary for the requested edit,
which can also result in unwanted artifacts in the final image.
Therefore, the best value for each edit instance and method is expected to vary. In Section~\ref{ssec:cfg-motivation} we will support this intuition with experiments.

\vspace{-1mm}
\subsection{Inversion-based editing models}
\vspace{-1mm}

Inversion-based editing models enable pretrained image generation models to edit existing images by first mapping the source image back into a latent noise representation and then regenerating it conditioned on an editing instruction. DDIM inversion \cite{Song2020-ng} introduced this idea. Later methods combined it with cross-attention control for text-conditioned editing \cite{hertz2023prompttoprompt}.
DDPM inversion \cite{huberman2024editfriendly} identified limitations of DDIM inversion and introduced a latent representation based on multiple noise maps that improves the editing results. 

\vspace{-1mm}
\subsection{Inversion-Free editing models}
\vspace{-1mm}

Inversion-free editing methods avoid mapping the source image back into a noise or latent representation. Instead, they construct a direct path from source to target distribution. This is natural for flow-matching models\cite{lipman2023flowmatching}, which learn a velocity field between noise and data.
FlowEdit\cite{kulikov2025flowedit} follows this approach, at each timestep it draws a shared noise sample to construct two points at the same noise level, one from noising the source image and one obtained by noising the current edited latent, then evaluates the velocity field at each point, conditioned on the source prompt and the target prompt respectively. The difference between these two predicted velocity fields is used to update the current latent directly from the source image toward the edited image distribution.

\section{Effect of instance-
specific CFG on image editing}
\label{ssec:cfg-motivation}

\begin{figure*}[t]
\centering
\begin{minipage}{0.38\textwidth}
\centering
\textbf{FlowEdit, SD3}\\[2pt]
\vspace{-1mm}
\includegraphics[width=\linewidth]{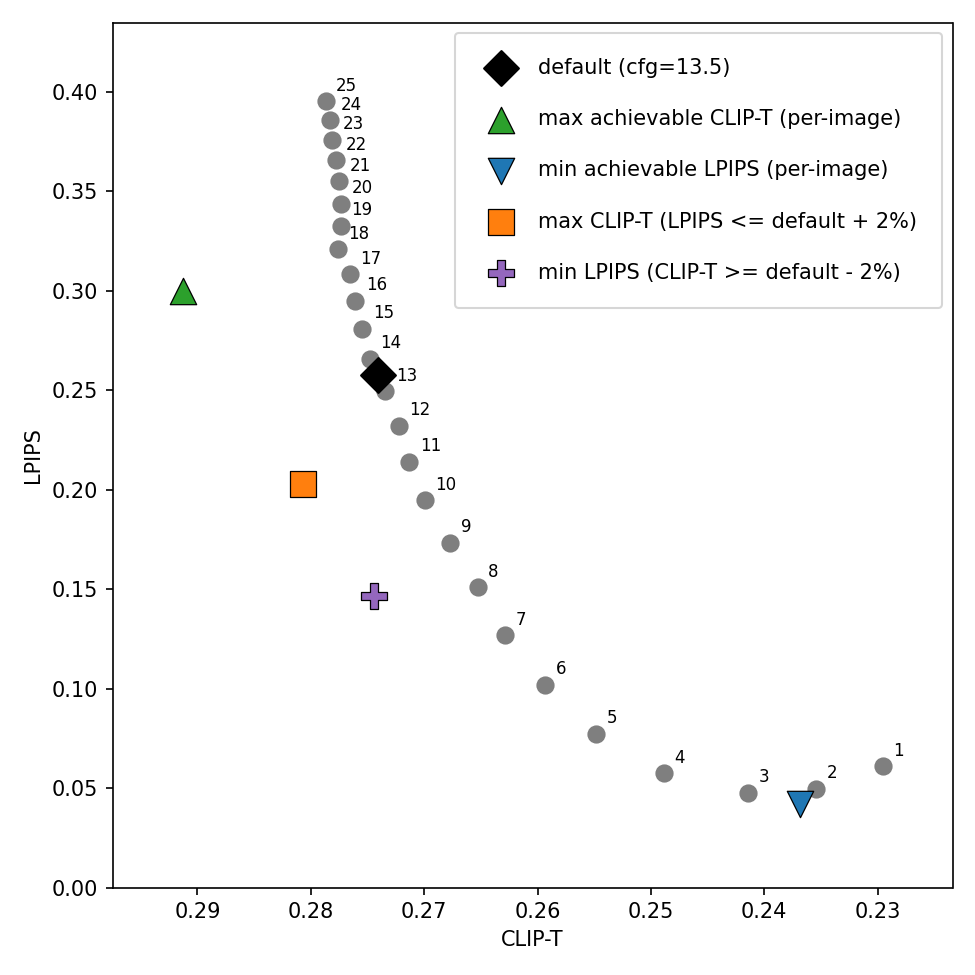}
\end{minipage}
\hspace{10mm}
\begin{minipage}{0.38\textwidth}
\centering
\textbf{DDPM-Inversion, SD1.4}\\[2pt]
\vspace{-1mm}
\includegraphics[width=\linewidth]{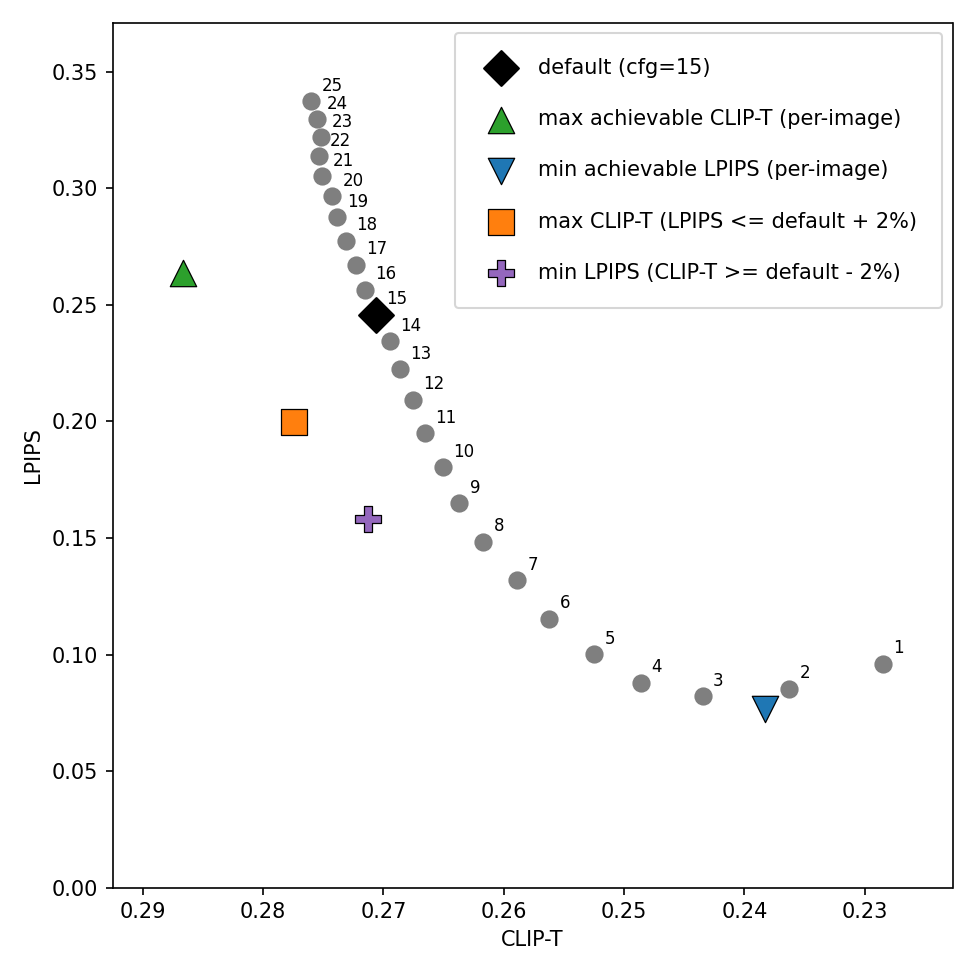}
\end{minipage}
\vspace{-3mm}
\caption{Average CLIP-T vs.\ LPIPS on PIE-Bench across 25 fixed CFG scales in $[1,25]$ (circles), with the default scale (diamond). Per-instance optimal CFG scale: max CLIP-T (up triangle), min LPIPS (down triangle), max CLIP-T within $\le2\%$ LPIPS degradation with respect to baseline (square), min LPIPS within $\le2\%$ CLIP-T degradation (plus).}
\label{fig:cfg-motivation}
\end{figure*}

In this section, we show the effect of CFG scale on the final editing result based on the metrics CLIP-T and LPIPS. 
We use the PIE-Bench dataset\cite{ju2023directinversion}: 700 images spanning a random-editing category and nine instruction-based categories (each split into artificial/natural subsets).
We run DDPM-Inversion\cite{huberman2024editfriendly} with SD1.4 \cite{rombach2022high} and FlowEdit\cite{kulikov2025flowedit} with SD3 \cite{esser2024sd3}. 

First, 
for each method, we calculated the average CLIP-T and LPIPS for each CFG scale value, and plot the results in gray circles in Fig.~\ref{fig:cfg-motivation}. Note that the x-axis is plotted in descending order.
The performance of the method's default CFG scale is marked with a black diamond.
As the CFG scale increases, the CLIP-T score increases, which implies that the edited images are more aligned with the target prompt on average,
and the LPIPS score also increases, which means the edited images are less similar to the source image on average. This trade-off behavior is expected. 
The question is: %
Can we push this Pareto frontier (and dominate some regions of it) by using instance-specific CFG scales?

To answer this question, we examine four tuning strategies. Two of them are obtained by identifying for each image the CFG scale that maximizes CLIP-T and the one that minimizes LPIPS, without any constraint on the other metric. The other two are obtained by identifying the CFG scale that maximizes CLIP-T while keeping LPIPS score within 2\% of the value obtained for the method's default CFG, and the CFG scale that minimizes LPIPS while keeping the CLIP-T score within 2\% of its default value. We average CLIP-T and LPIPS across all image instances for each of the four cases, and add them to Fig.~\ref{fig:cfg-motivation}.

It can be observed that all the instance-wise tuning strategies that account for the CLIP-T metric dominate significant regions of the Pareto frontier (i.e., outperform points in the frontier in both metrics).
Since a user may be interested in gains compared to the default setup, let us focus on the strategy that caps the CLIP-T degradation to 2\% of its default value,
marked in a purple plus symbol.
Inspecting the performance of this point, we see that the change in CLIP-T score is negligible while 
the LPIPS score improves by $\sim$0.1 in both methods.
This significant improvement in fidelity to the source image without compromising on realizing the edit is a key finding of our work. It motivates our next step:
We want to find a way to select the CFG scale per edit instance in an efficient way, without repeated activations of editing method.

\section{Efficient tuning method}
\label{sec:method}

We propose a lightweight 4-hidden-layer MLP with 3.8 million parameters for each metric (note that this size is a negligible compared to the scale of the diffusion/flow models). The MLP takes as input the CLIP embedding of the source image, source and
target prompt, the metrics for the method's default CFG scale, and the candidate CFG scale. Per editing method and underlying diffusion/flow model, we train this model on a large dataset made of images edited by the method with different CFG scale values, where metrics of the edited images are computed and used as labels. 
At inference time, given an editing instance, we use the trained metric prediction models to select the CFG scale that optimizes the estimated target metric (LPIPS) capping the other estimated metric (CLIP-T) by the performance of the method's default setting.

\subsection{Training and evaluating the metric predictor}
\label{ssec:setup}

We use UltraEdit\cite{zhao2024ultraedit} and EmuEdit\cite{sheynin2023emuedit} for training, validation and testing of the metric predictor. UltraEdit is a large-scale, automatically generated instruction-based image editing dataset with $\sim$4M samples across nine instructions. We use the first 52k images and sample several random CFG scales per image. %
EmuEdit is a smaller instruction-based editing dataset, with 5,611 publicly available images, of which 4,452 have valid prompts, we use these and sample 10 CFG scale values per image. %
Training is done using the SGD optimizaer with L1 loss for 200 epochs, with patience for 30 epochs as early stopping criterion, a learning rate of 0.01, momentum 0.9 and weight-decay of $1 \times 10^{-4}$.

We evaluate the accuracy of the metric predictors separately per metric and editing method, on a held-out split of UltraEdit and EmuEdit (in-distribution) and on PIE-Bench (out-of-distribution).
We report four complementary metrics. The mean absolute error (MAE). 
The coefficient of determination $R^2$ between the predicted and ground-truth metrics.
The Pearson correlation $r$ and per-image Spearman rank correlation $\rho$.
The results are reported in Tables \ref{tab:predictor-accuracy-id} and \ref{tab:predictor-accuracy-ood}.

\begingroup
\setlength{\intextsep}{4pt}
\begin{table}[t]
\centering
\caption{In-distribution predictor accuracy (held-out test split). Per-image $\rho$ uses 3{,}548 of 8{,}468 test images with $\geq$3 CFG samples (5-9 each), images with a single CFG sample cannot support a ranking.}
\label{tab:predictor-accuracy-id}
\small
\setlength{\tabcolsep}{3pt}
\begin{tabular}{llcccc}
\hline
Model & Target & MAE & $R^2$ & Pearson & $\rho$ (per-img.) \\
\hline
\multirow{2}{*}{FlowEdit} & LPIPS & 0.0160 & 0.971 & 0.985 & 0.985 \\
 & CLIP-T & 0.0140 & 0.825 & 0.909 & 0.622 \\
\multirow{2}{*}{DDPM-Inversion} & LPIPS & 0.0184 & 0.941 & 0.970 & 0.972 \\
 & CLIP-T & 0.0146 & 0.793 & 0.891 & 0.610 \\
\hline
\end{tabular}

\vspace{3mm}

\centering
\caption{Out-of-distribution predictor accuracy on PIE-Bench (700 images $\times$ 20 CFG scales = 14{,}000 pairs).}
\label{tab:predictor-accuracy-ood}
\small
\setlength{\tabcolsep}{3pt}
\begin{tabular}{llcccc}
\hline
Model & Target & MAE & $R^2$ & Pearson & $\rho$ (per-img.) \\
\hline
\multirow{2}{*}{FlowEdit} & LPIPS & 0.022 & 0.947 & 0.978 & 0.992 \\
 & CLIP-T & 0.0125 & 0.822 & 0.908 & 0.543 \\
\multirow{2}{*}{DDPM-Inversion} & LPIPS & 0.0207 & 0.929 & 0.965 & 0.984 \\
 & CLIP-T & 0.0125 & 0.802 & 0.899 & 0.569 \\
\hline
\end{tabular}
\end{table}
\endgroup
LPIPS is consistently easier to predict than CLIP-T. $R^2$ is higher for LPIPS in both methods and both distributions, and per-image rank agreement $\rho$ across CFG candidates is much stronger. %
The limited predictability of CLIP-T is the main barrier to matching exhaustive search. The gap between consecutive CFG scales on this axis is small, so a more accurate predictor is needed for reliable per-image ranking and CFG selection. We will analyze the core hardness of predicting the CLIP-T metric and focus on improving it in an extended version of this paper.

\vspace{-1mm}
\subsection{Editing with predictor-based CFG-selection}
\label{ssec:selection-quality}
\vspace{-1mm}

For choosing the CFG scale per edit instance, using our predictor, we use a rule similar to that in Section \ref{ssec:cfg-motivation}: minimize estimated LPIPS while keeping the estimated CLIP-T score within 2\% of its value for the baseline's default CFG. %

\begin{table}[t]
\centering
\small
\caption{Average per-image runtime, measured on 50 PIE-Bench images with NVIDIA RTX 6000 Ada GPU CUDA-13.}
\label{tab:efficiency}
\begin{tabular}{lccc}
\hline
 & Baseline (s) & Exhaustive (s) & Ours (s) \\
\hline
FlowEdit & 6.8 & 147.13 & 14.4 \\
DDPM-Inversion & 17.5 & 154.38 & 24.2\\
\hline
\end{tabular}

\vspace{3mm}

\centering
\small
\caption{Average CLIP-T and LPIPS on PIE-Bench: ground-truth optimal CFG scale vs. our predicted per-image CFG scale using a $2\%$ CLIP-T degradation threshold.}
\label{tab:selection-quality}
\setlength{\tabcolsep}{3pt}
\begin{tabular}{llcc}
\hline
Model & Selection & CLIP-T & LPIPS \\
\hline
\multirow{2}{*}{FlowEdit} 
& Baseline & 0.2741 & 0.2577 \\
& Exhaustive  & 0.2746 & 0.1469 \\
 & Ours (predicted CFG) & 0.2704 & 0.1731 \\
\multirow{2}{*}{DDPM-Inversion} 
& Baseline & 0.2706 & 0.2458 \\
& Exhaustive  & 0.2712 & 0.1580 \\
 & Ours (predicted CFG) & 0.2663 & 0.1702 \\
\hline
\end{tabular}
\vspace{-2mm}
\end{table}

The advantage of our approach over the exhaustive search is requiring at most two executions of the editing method instead of one execution per CFG scale candidate. 
To quantify this, we measure the average per-image runtime of both approaches on 50 images with 20 CFG scale candidates, for DDPM-Inversion and FlowEdit, and report the results in Table~\ref{tab:efficiency}.
Our pipline's runtime includes one execution of the editing method with the default CFG scale (i.e., the baseline setting), 20 predictor calls for LPIPS and 20 for CLIP-T ($\sim$ 1ms per prediction), and one execution of the editing method with the selected CFG scale. The exhaustive search's runtime includes 20 executions of the editing method---one for each CFG scale candidate---which is significantly slower.

Table~\ref{tab:selection-quality} compares editing with our predicted per-instance CFG selection against the exhaustive-search performance bounds and the fixed default-CFG baseline.
For FlowEdit, our approach achieves a 1.35\% worse CLIP-T than the default and $\sim$33\% better LPIPS.
For DDPM-Inversion, we achieve a 1.59\% drop in CLIP-T for $\sim$31\% improvement in LPIPS over the baseline.
Fig.~\ref{fig:qualitative-examples} shows qualitative examples of edits produced with our selected CFG scale alongside the baseline, illustrating that our approach yields edits aligned with the target prompt while better preserving the source image.

\begingroup
\renewcommand{\gridw}{1.02in}
\begin{figure}[t]
\centering
\footnotesize\textbf{DDPM-Inversion, SD1.4}\\[1pt]
\subcaptionbox{Source image}[\gridw]{\includegraphics[width=\gridw]{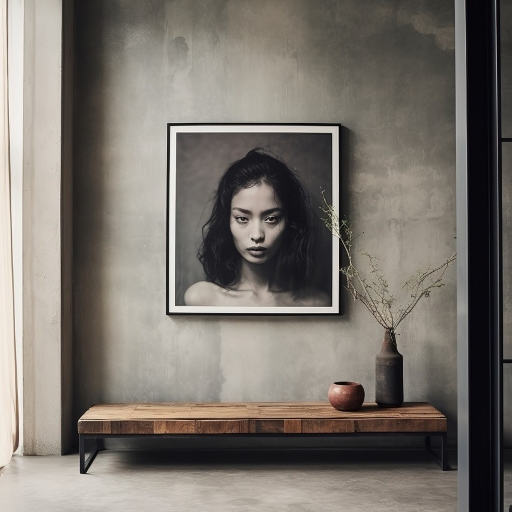}}\hfill
\subcaptionbox{Baseline (default CFG 15)}[\gridw]{\includegraphics[width=\gridw]{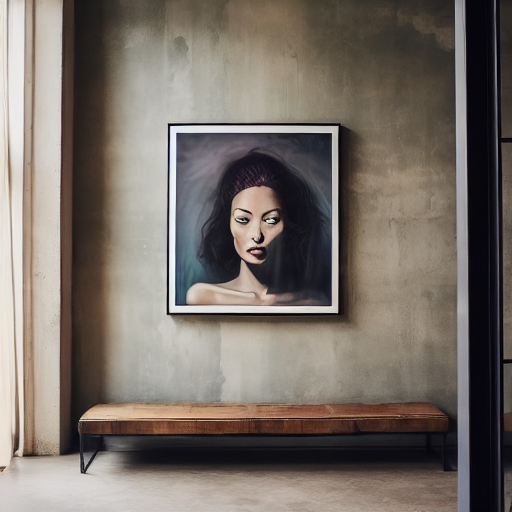}}\hfill
\subcaptionbox{Ours (predicted CFG 10)}[\gridw]{\includegraphics[width=\gridw]{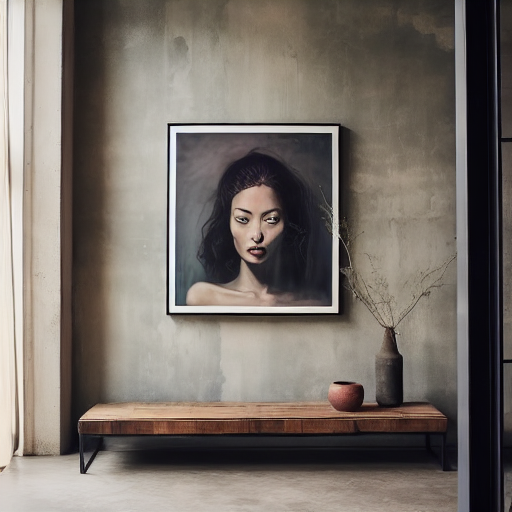}}\\[2pt]
{\scriptsize\itshape Prompt: Change the photo on the wall to a painting}

\vspace{4pt}
\footnotesize\textbf{FlowEdit, SD3}\\[1pt]
\subcaptionbox{Source image}[\gridw]{\includegraphics[width=\gridw]{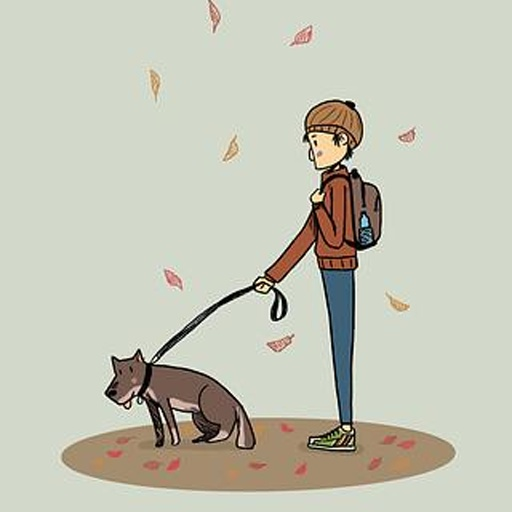}}\hfill
\subcaptionbox{Baseline (default CFG 13.5)}[\gridw]{\includegraphics[width=\gridw]{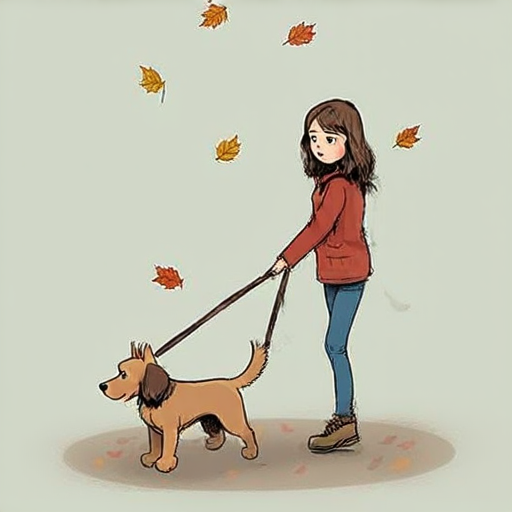}}\hfill
\subcaptionbox{Ours (predicted CFG 4)}[\gridw]{\includegraphics[width=\gridw]{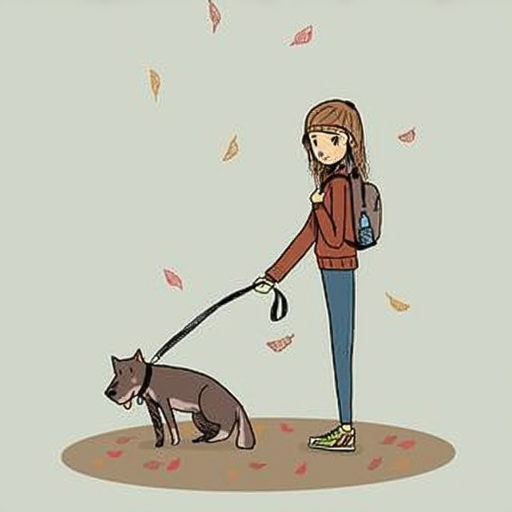}}\\[2pt]
{\scriptsize\itshape Prompt: Change the gender of the person walking their dog}
\vspace{-1mm}
\caption{Qualitative comparison. From left to right: source image, default-CFG edit, and our predicted-CFG edit.}
\label{fig:qualitative-examples}
\end{figure}
\endgroup

\vspace{-1mm}
\section{Conclusion}
\vspace{-1mm}

We showed that instance-specific CFG scale selection can significantly improve image fidelity with negligible effect on edit realization, compared to a fixed default scale. 
To reduce the cost of tuning at inference time,
we trained a lightweight neural network to predicts the editing method's metrics for a given editing instance and CFG scale in negligible time. %
Using it, we proposed an efficient procedure for editing with instance-specific CFG scale value that
approaches the editing performance of exhaustive search.

\vfill\pagebreak

\small 
\bibliographystyle{IEEEbib}
\let\oldthebibliography\thebibliography
\renewcommand{\thebibliography}[1]{%
  \oldthebibliography{#1}%
  \setlength{\itemsep}{0pt}%
  \setlength{\parskip}{0pt}%
}
\bibliography{strings,refs}

\end{document}